\documentclass{article}
\usepackage[margin=1in]{geometry}
\usepackage{graphicx} 
\usepackage{hyperref}
\usepackage{amsmath}
\usepackage{xcolor}
\usepackage{xspace}
\usepackage{cite}
\usepackage{booktabs}
\usepackage[affil-it]{authblk}

\usepackage[normalem]{ulem} 

\newcommand{\TTCfull}{\ensuremath{C_2(q, t_1, t_2)}\xspace}
\newcommand{\gTwofull}{\ensuremath{g_2(q, \tau)}\xspace}
\newcommand{\TTC}{\ensuremath{C_2}\xspace}
\newcommand{\TTCraw}{\ensuremath{C_{2,\text{raw}}}\xspace}
\newcommand{\TTCden}{\ensuremath{C_{2,\text{denoised}}}\xspace}
\newcommand{\It}{\ensuremath{I(q, t)}\xspace}
\newcommand{\gTwo}{\ensuremath{g_2}\xspace}
\newcommand{\gOne}{\ensuremath{g_1(q, \tau)}\xspace}
\newcommand{\lag}{\ensuremath{\tau}\xspace}
\newcommand{\tage}{\ensuremath{t_{\text{age}}}\xspace}
\newcommand{\Cinf}{\ensuremath{C_{\infty}}\xspace}
\newcommand{\bobs}{\ensuremath{\beta_{\text{obs}}}\xspace}

\title{A Fully Convolutional Approach to Denoising 2D Correlation Spectra}

\author{Nisar Nellikunnummel\thanks{\texttt{nnellikun@bnl.gov}}}
\author{Andi M Barbour\thanks{\texttt{abarbour@bnl.gov}}}
\author{Lutz Wiegart}
\author{Tatiana Konstantinova\thanks{Currently at: Amazon}}
\author{Anthony M DeGennaro\thanks{Currently at: GE Aerospace Research}}

\affil{Brookhaven National Laboratory, Upton, NY 11973, USA}

\date{July 28, 2026}

\begin{document}

\maketitle
\begin{abstract}

We present a fully convolutional denoising autoencoder (FC-DAE) tailored for two-dimensional representations of dynamic correlations that is applicable to many experimental techniques. Here, we demonstrate its performance on two-time intensity correlation functions (\TTC) from X-ray photon correlation spectroscopy (XPCS).
Unlike conventional denoising autoencoders that are typically restricted to fixed input sizes, the FC-DAE accepts inputs of arbitrary dimensions while preserving correlation structures across diverse dynamical regimes. The model is trained using experimentally derived \TTC data collected at NSLS-II beamlines, with data augmentation applied to expand the diversity of the dataset and reduce overfitting. The FC-DAE successfully recovers intricate dynamical features in low signal-to-noise conditions while maintaining structural fidelity. To assess reconstruction reliability, we employ quantitative metrics to evaluate structural fidelity and identify potential model-induced bias. Our results demonstrate that the FC-DAE provides robust denoising performance with high computational efficiency, enabling recovery of XPCS dynamics under photon-limited and low-dose measurement conditions.
\end{abstract}

\section{Introduction}
\label{sec:intro}

Two-dimensional correlation analysis is widely used across the physical sciences to quantify non-equilibrium phenomena, track system memory, and isolate dynamic fluctuations from noise. 
These second-order statistical methods are fundamental to X-ray photon correlation spectroscopy (XPCS), but are also applied in other techniques, such as electron microscopy~\cite{zhang2018spatially, spangenberg2021direct, huang2024momentum}.
XPCS probes structural dynamics in materials by measuring intensity fluctuations in speckle pattern obtained by coherent scattering from internal inhomogeneities in electron density~\cite{Shpyrko2014XPCS}. With respect to coherent X-ray photon probes, the technique is well-suited for investigating both equilibrium and non-equilibrium dynamics across a wide range of systems. Specifically, XPCS is used to probe mesoscale dynamics in processes such as the aging of glasses~\cite{Cipelletti2000Glassy}, colloidal gelation~\cite{Fluerasu2007Colloidal}, 
polymer relaxation~\cite{Burghardt2007Polymer}, or the fluctuations of charge-orbital order in magnetite~\cite{hua2023discerning} as well as more complex processes like egg yolk heating~\cite{anthuparambil2023exploring} and ferroelectric domain transformations~\cite{liu2025dynamics}. The two-time intensity correlation function, \TTCfull, is used to quantify the temporal evolution of speckle intensity fluctuations in the scattered X-ray signal. It is defined as:

\begin{equation}
\TTCfull = \frac{\langle I(q, t_1) I(q, t_2) \rangle}{\langle I(q, t_1) \rangle  \langle I(q, t_2) \rangle}
\label{C2_def}
\end{equation}

where \It represents the scattered intensity at wave vector $q$ and time $t$, and $<>$ denotes an ensemble average, typically performed over a range of detector pixels at constant $q$. \TTCfull is a two-dimensional matrix that quantifies the correlation between intensities at times $t_1$ and $t_2$. The \TTCfull matrix is symmetric about its main diagonal ($t_1=t_2$), allowing the raw time coordinates $t_1$ and $t_2$ to be transformed into physically meaningful axes: the age axis, \tage, and the lag axis, \lag~\cite{Madsen_2010}.
The age axis represents the evolution of the system since the start of the measurement and is defined as the average of the two time points, $\tage = (t_1 + t_2) / 2$. The lag axis represents the duration of the measured fluctuations and is defined as the time difference, $\lag = |t_2 - t_1|$. In the limit of stationary dynamics, \TTCfull reduces to the one-time correlation function, \gTwofull, by averaging \TTCfull over the \tage axis at a constant \lag.

\begin{equation}
\gTwofull = \langle \TTCfull \rangle_{\tage}
\label{g2_avg}
\end{equation}

Henceforth, \TTCfull\ is denoted as \TTC, and \gTwofull\ as \gTwo. The time variables $t_1$, $t_2$, \tage, and \lag\ are discretized in units of detector frames, corresponding to uniformly sampled time intervals (lags) during XPCS measurements. Thus, all correlation functions are evaluated on a discrete time grid, where the frame index serves as a proxy for physical time (lag time) and can be converted to real time using the known frame acquisition rate. 

The distinction between equilibrium and nonequilibrium dynamics lies in whether relaxation is time-invariant. In equilibrium, \TTC depends only on lag time, \lag, leading to a constant relaxation rate and uniform correlation width along the age axis \tage\ in the \TTC. In contrast, nonequilibrium dynamics are non-stationary, with relaxation times evolving over \tage, appearing as broadening or narrowing of the correlation profile. They may also include stochastic events, such as avalanches or sudden structural rearrangements, which manifest as discrete breaks or discontinuous dynamics along \tage.

To interpret \gTwo physically, it is related to the underlying dynamics through the Siegert relation~\cite{sutton2008review}:
\begin{equation}
    \gTwo = \Cinf + \beta |\gOne|^2 
\end{equation}
where $\Cinf$ denotes the baseline and $\beta_0=\beta+\Cinf-1$ ($0 < \beta_0 \le 1$) is the instrumental coherence factor, or contrast. The normalized intermediate scattering function, \gOne, encodes the sample dynamics, with different functional forms corresponding to different relaxation behaviors. Consequently, the measured \gTwo is fitted to models appropriate for the dynamics under study. In nonequilibrium systems, where the dynamics evolve with \tage, this relation is applied locally within quasi-stationary regions of the data.

In practice, nonequilibrium XPCS analysis often requires slicing \TTC along the age axis \tage to obtain locally stationary regions suitable for fitting. For noisy data, adjacent slices are typically averaged over an age window $\Delta\tage$ to achieve sufficient signal-to-noise ratio (SNR) for reliable parameter extraction. This approach implicitly assumes that the dynamics remain approximately stationary within $\Delta\tage$; however, when this condition is not satisfied, the averaging process reduces temporal resolution and can introduce bias in the extracted parameters. Furthermore, selecting an appropriate binning window often requires expert judgment and becomes increasingly impractical for large datasets.

These challenges are further compounded by the inherently noisy nature of experimental \TTC data. In practice, the dominant limitations arise not only from detector imperfections but, more fundamentally, from photon-limited measurement conditions. As a result, effective denoising methods are essential for improving data quality and enabling reliable extraction of dynamical information. The primary challenges and the role of denoising in addressing them are summarized as follows:

\begin{itemize}
    \item \textbf{Photon-limited conditions:} Limited coherent flux reduces the number of detected photons per frame, such that increasing temporal resolution or probing larger $q$ (smaller length scales) leads to reduced SNR. Denoising enables recovery of meaningful correlation signals under these conditions.
    
    \item \textbf{Low-dose measurements:} Radiation-sensitive samples often require reduced X-ray exposure to mitigate beam damage, which further degrades SNR. Denoising allows reliable extraction of dynamics while operating at lower dose.
    
    \item \textbf{Detector artifacts:} Occasional detector-related events, such as hot pixels, can introduce temporally correlated artifacts in \TTC (e.g., streaks along rows and columns). Other times, constantly evolving correlated noise of the detector is unavoidably correlated, resulting in may more streaks. Denoising helps suppress these distortions and improves overall data fidelity.
\end{itemize}

Conventional denoising methods, such as Gaussian and mean filtering~\cite{GonzalezWoods2008,Al-amri2010NoiseRemoval}, suppress noise by averaging neighboring values, often blurring fine correlation features essential for accurate dynamical analysis. Denoising Autoencoders (DAEs) \cite{Konstantinova2021, Konstantinova2022SpeckleML, Timmermann2022} offer a promising solution by learning the functional form of the dynamic signal obscured by noisy data, enabling effective denoising while preserving the intrinsic dynamics. However, a key limitation of DAEs is their reliance on a fixed latent space, which requires inputs of a fixed size and restricts the model's flexibility when handling arbitrarily sized data. Additionally, the use of a fixed latent representation introduces significant bias, even when the latent space is relatively large. While DAEs can effectively denoise complex patterns present in the training domain, they face a significant challenge in generalization. 
Because the latent space of a DAE is shaped by the dynamics represented in the training data~\cite{Konstantinova2021,Timmermann2022}, the model may struggle to generalize to qualitatively different correlation patterns.
This highlights two major demerits of the DAE approach: first, the necessity for massive, high-fidelity datasets to populate the latent space; and second, the difficulty of training a single 'universal' model. Instead, DAEs often require specialized models tailored to specific classes of dynamics, as any signal not explicitly captured during the large-scale training phase is likely to be misinterpreted as noise and suppressed.

To overcome these limitations, we propose a Fully Convolutional Denoising Autoencoder (FC-DAE) based on a Fully Convolutional Network architecture~\cite{FC-DAE_original}.
By eliminating fully connected layers, the FC-DAE can process \TTC{}s of arbitrary spatial dimensions, making it directly applicable to experimental XPCS datasets without the need for resizing or cropping~\cite{Horwath2024AINERD}. More importantly, the model preserves physically meaningful structures in noisy data and can capture complex oscillatory or intermittent features beyond dynamics well described by the standard Kohlrausch–Williams–Watts (KWW)~\cite{Williams1970} function.
In addition to the architectural advancement, we introduce a set of quantitative reliability metrics to assess the fidelity and applicability of the denoised outputs, addressing the critical need to evaluate model-induced bias in scientific analysis. Fig.~\ref{C2_types} illustrates representative cases where the latent-space DAE fails to recover underlying correlation features, either by over-smoothing or distorting the signal, while the FC-DAE retains these structures with significantly improved clarity. In this work, we demonstrate the effectiveness of the proposed approach using experimental XPCS datasets spanning a wide range of dynamical regimes, and show that it enables more reliable extraction of dynamic parameters from noisy measurements.

\section{Dataset Preparation and Preprocessing}

\subsection{Experimental Dataset}

\begin{figure}
\centering
\includegraphics[scale=0.35]{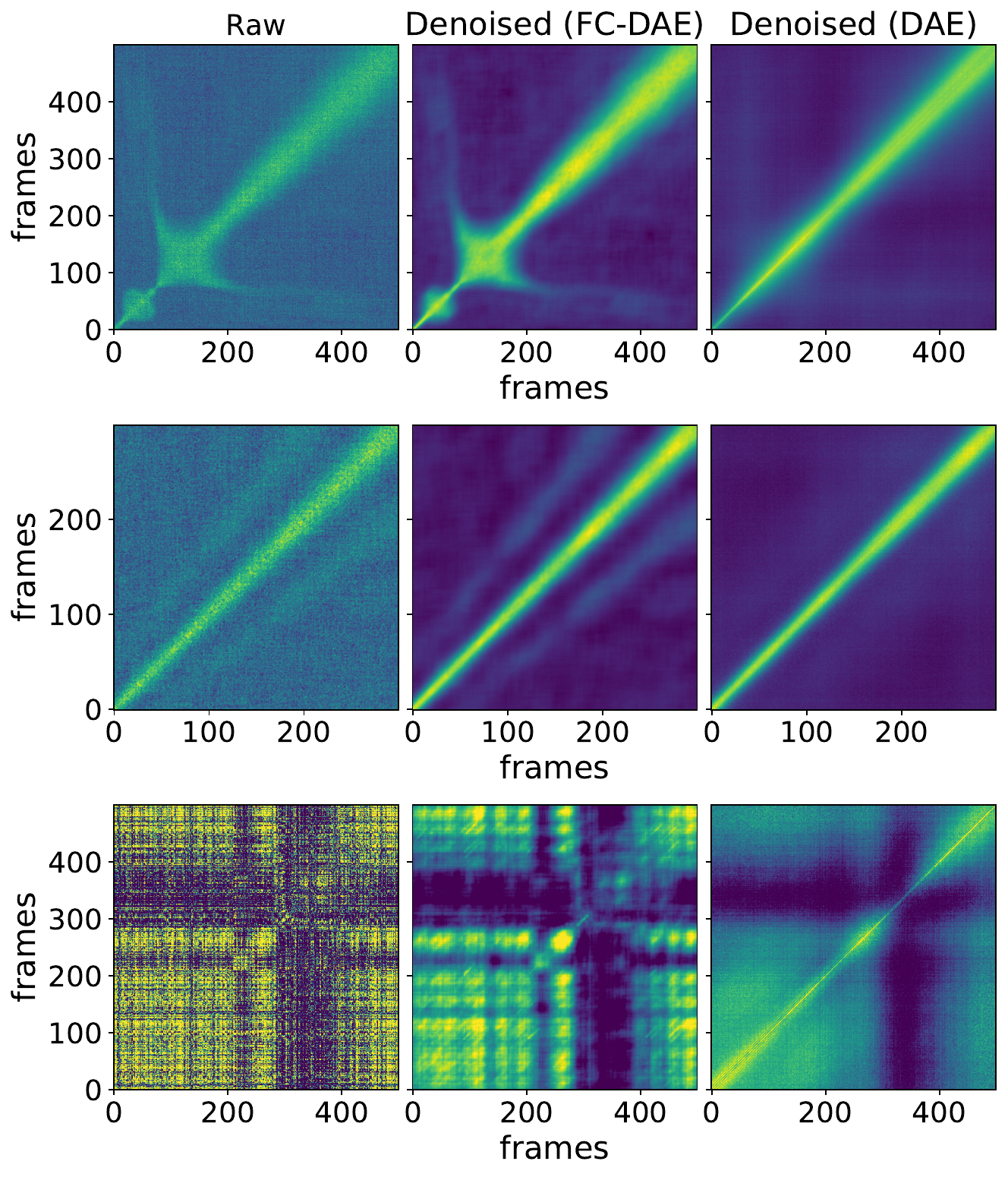}
\caption{Representative examples of raw and denoised \TTC. The raw \TTC (left) is compared with denoised results from the FC-DAE (middle) and the conventional DAE (right). The FC-DAE successfully suppresses noise and preserves structural features where the DAE fails.}
\label{C2_types}
\end{figure}

Model training is conducted using data collected at the Coherent Hard X-ray Scattering (CHX) and Coherent Soft X-ray Scattering (CSX) beamlines of NSLS-II. In order for the model to denoise experimental data exhibiting a wide range of dynamical behaviors, the training dataset includes diverse \TTC patterns collected from XPCS experiments.
Representative examples of such \TTC patterns are shown in Fig.~\ref{C2_types}. Many of these patterns are not adequately handled by previous DAE models \cite{Konstantinova2022SpeckleML, Konstantinova2021}.
We used data from 118 unique experiments and selected several regions of interest in reciprocal space for each experiment when calculating \TTC. In total, 779 \TTC patterns were used for training, 56 for testing, and 68 for validation. The size of the \TTC patterns ranges from 134 to 2995 frames. 
Unlike previous DAE models trained primarily on datasets describable by KWW dynamics within individual $\Delta\tage$ bins, our training dataset includes a broader range of complex and non-KWW-like dynamical behaviors.

\subsection{Target Generation and Preprocessing}
\label{subsec:dat_processing}

Generating experimental ground-truth pairs (noisy, noise-free) across a broad range of dynamics, timescales, and noise conditions is often unfeasible. To address this, we employ the DAE architecture of \cite{Konstantinova2022SpeckleML} as a target preparation model. We use raw \TTC (\TTCraw) as noisy inputs and the corresponding DAE outputs as proxy ‘noise-free’ targets, as illustrated by the top portion of Fig.~\ref{fig:FC-DAE}. Although the DAE is not capable of accurately reconstructing large \TTC	maps with complex nonstationary dynamics, its performance remains highly reliable when restricted to localized subregions (such as near-diagonal areas characterized by simpler, quasi-equilibrium dynamics). In our approach, we exploit this by using the DAE exclusively in these small subregions to generate stable, physically consistent surrogate targets for training. Importantly, the FC-DAE is not constrained by the limitations of the DAE. By utilizing a fully convolutional architecture, the model inherits a spatial invariance that forces local smoothing kernels to function uniformly across the entire map. Consequently, by training on targets that are already partially smoothed in reliable local regions, the FC-DAE learns robust denoising strategies that seamlessly generalize to larger scales and more complex global structures, ultimately outperforming its teacher model.

The DAE accepts only input data of a fixed size ($100\times100$), and inputs are generated by cropping along the diagonal of the \TTCraw. A copy of each \TTCraw with reversed \tage is also included to increase variability. Additional data augmentation is performed by subsampling frames at different intervals—every 2nd, 3rd, 4th, 10th, 20th, and 30th frame—to generate new \TTCraw. 
The diagonal elements of \TTCraw parallel to \tage represent trivial self-correlations dominated by photon counting statistics and instrumental noise rather than true dynamics. Therefore, prior to cropping, these values are replaced by the average of their nearest off-diagonal neighbors.

To assess robustness of model under varying signal-to-noise conditions, we generate bootstrapped \TTCraw samples by randomly subsampling pixels within each frame prior to correlation calculation. Subsampling fractions of 50\%, 25\%, 10\%, and 5\% are used to systematically increase noise levels, producing multiple noisy realizations of the same underlying dynamics for validation with known, high fidelity ground truths.

Each input \TTCraw is standardized to have zero mean and unit standard deviation before training. Standardizing the data ensures consistent feature scales for stable autoencoder training, enabling more effective noise removal by focusing the model on dynamic patterns rather than absolute signal magnitudes. Consequently, during model application, input data is first standardized. After the model generates the denoised output, the standardization is reversed to bring this output back to the original data's scale.

\section{FC-DAE Model Architecture}

\begin{figure}
\centering
\includegraphics[scale=0.35]{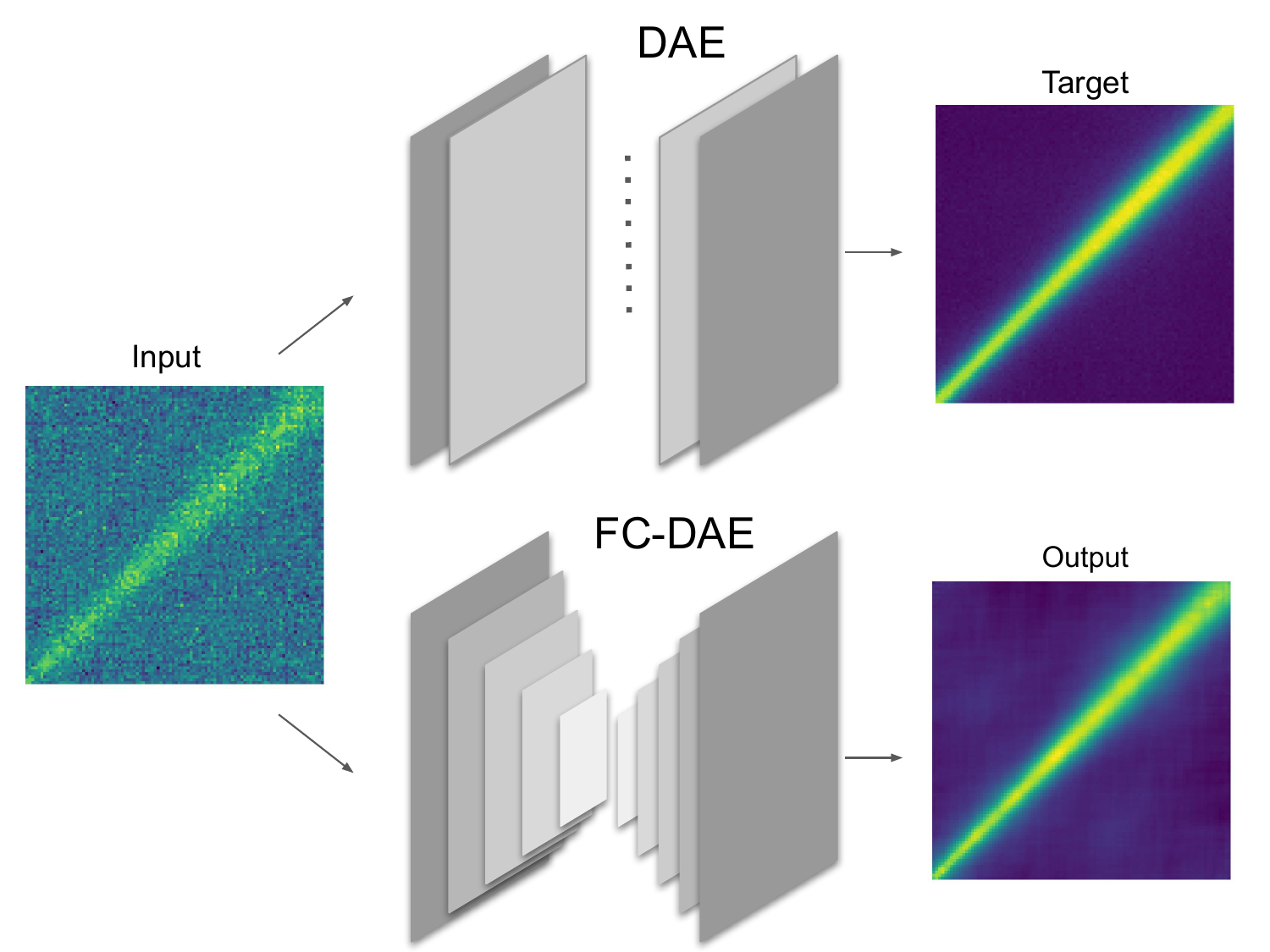}
\caption{Architecture of the FC-DAE in a supervised training setup. The FC-DAE comprises multiple convolutional layers designed to denoise input images. During training, the output image is compared to a target (clean) image using a loss function, and the convolutional filters are optimized accordingly.}
\label{fig:FC-DAE}
\end{figure}

The model architecture is composed entirely of convolutional layers, structured into an encoder–decoder framework as illustrated in the lower portion of Fig~\ref{fig:FC-DAE}. The encoder consists of five convolutional layers with 1, 4, 8, 16 and 32 output channels, respectively.
The decoder mirrors this structure using transposed convolutional layers to reconstruct the denoised output. A kernel size of 3 is used consistently across all layers. Batch normalization is applied after each convolutional layer (or transposed convolutional) and before the activation function. The exponential linear unit (ELU) activation is used throughout the network, except in the final output layer, where no activation is applied. Strides of size 1 are used in each layer and the output of each layer is padded to preserve the spatial dimensions of the input.

We also investigated deeper variants of the encoder-decoder architecture by extending the five-layer model to six and seven convolutional layers, introducing additional encoder layers with 64 and 128 output channels, respectively.
While these deeper configurations significantly increased the number of trainable parameters and computational cost, no measurable improvement in denoising performance was observed compared with the baseline five-layer architecture. In addition, we explored the use of a larger convolutional kernel size of $5\times5$ in all layers to increase the effective receptive field. This modification likewise resulted in increased computational demand without yielding any noticeable performance gains. Consequently, the five-layer architecture with a $3\times3$ kernel was selected as the optimal trade-off between model complexity, computational efficiency, and denoising performance.

\section{Training Strategy}

The model learns convolutional filters tailored to the characteristics of the input data through supervised training. Training data preparation is described in Section~\ref{subsec:dat_processing}.
The mean square error (MSE) between the output and the target is used as the training loss function. The convolutional kernel size controls the balance between capturing fine local details and broader contextual information, thereby influencing the trade-off between noise removal and detail preservation. To prevent overfitting, early stopping is applied during training when the loss function reaches a predefined threshold. Model optimization is performed using the Adam optimizer~\cite{kingma2017adammethodstochasticoptimization}.

\section{Model Variance}

To quantify model uncertainty arising from random weight initialization, we estimated the variance of the FC-DAE using an ensemble approach. 
Ten independent models were trained with identical hyperparameters but different initialization seeds, ensuring distinct initial conditions while maintaining reproducibility.
The validation set was used to assess the stability of the ensemble. For each validation sample, denoised outputs from the ten models were compared pixel-by-pixel across the \TTCden maps to compute the ensemble variance, which was then averaged to obtain a single scalar variance per sample.
Across all validation samples, the ensemble variance remained extremely low ($1.2\text{--}1.7 \times 10^{-4}$
), indicating highly consistent predictions across different initializations. 
This demonstrates that the learned denoising mapping is stable and that uncertainty due to random initialization is negligible compared to intrinsic noise in the raw inputs. These promising results support the reliable quantitative extraction of dynamics from denoised \TTC{}s.

To assess the impact of model uncertainty on the extracted dynamics, we compared the ensemble variance of the autoencoder predictions with the corresponding observed contrast $\bobs$ (estimated from the statistical distribution of the \TTC map using the $1^{\text{st}}$ and $99^{\text{th}}$ percentiles) for the validation sample. For each sample, ten independently initialized models were evaluated, and the pixel-wise variance across the ensemble was computed for each frame and averaged to obtain a single variance metric per sample, yielding an overall mean of $\langle \sigma_{\mathrm{ens}}^{2} \rangle = (1.457 \pm 0.016) \times 10^{-4}$. This is significantly lower than the observed mean contrast of $\langle \bobs \rangle = (1.84 \pm 0.12) \times 10^{-1}$. Consequently, the ratio $\sigma_{\mathrm{ens}}^{2}/\bobs$ typically falls on the order of $10^{-3}$, with a median value of $1.0 \times 10^{-3}$ and a 10–90 percentile range of $(4.6 \times 10^{-4}, 1.5 \times 10^{-3})$. These results demonstrate that the ensemble variance remains approximately three orders of magnitude smaller than \bobs, indicating that uncertainty due to random initialization has a negligible impact on the quantitative extraction of dynamical parameters.

\section{Model Scope and Reliability Metrics}
\label{sec:Scop_Reliable_Mertic}

From a practical perspective, the primary source of uncertainty in the model arises from intrinsic bias introduced during denoising. Any denoising autoencoder must impose some bias to suppress noise while preserving meaningful structural information. Consequently, assessing the reliability of the denoised output requires evaluating how faithfully it represents the underlying sample dynamics. Although the FC-DAE is designed to handle a broad range of \TTC patterns, including complex dynamical behaviors, some cases cannot be reliably reconstructed. To address this, we introduce a set of quantitative metrics that evaluate different aspects of reconstruction reliability and determine whether the denoiser can be confidently applied to a given dataset. These metrics are evaluated on a validation set spanning a wide range of dynamical behaviors, including both well- and poorly reconstructed cases.




\subsection{Relative Contrast Shift Analysis}

To quantify the systematic bias introduced by FC-DAE, we evaluate the relative shift in observed contrast, defined as $\Delta \beta_{\text{obs, rel}} = (\beta_{\text{obs, denoised}} - \beta_{\text{obs, raw}}) / \beta_{\text{obs, raw}}$,  where $\beta_{\text{obs, denoised}}$ represents the observed contrast after applying the denoising process and $\beta_{\text{obs, raw}}$ represents the original observed contrast prior to denoising.
In XPCS, the speckle contrast is a fundamental physical parameter; therefore, a high-fidelity denoiser must preserve it without introducing artificial inflation or suppression.

Analysis of validation samples reveals that in high-contrast cases ($\beta_{\text{obs, raw}} > 0.1$), the model achieves high fidelity, with relative errors typically remaining below 5\%.
Conversely, significant bias ($\Delta \beta_{\text{obs, rel}} > 20\%$) is observed exclusively in low-signal regimes where $\beta_{\text{obs, raw}} < 0.05$. In these cases, the noise floor (measured as the standard deviation of the uncorrelated baseline) becomes comparable to the signal itself, leading the model to occasionally misidentify noise fluctuations as part of the physical correlation ridge. Overall, the relative contrast shift quantifies the robustness of FC-DAE across different \TTC signal regimes.

\subsection{Residual Autocorrelation Analysis for Denoising Validation}

To evaluate whether the FC-DAE effectively removes noise without distorting the underlying signal of the measured \TTC, we analyze the residual—defined as the difference between the raw and denoised signals: $\TTCden-\TTCraw$. Ideally, if the model captures all physically meaningful structures in the \TTC map, the residual should consist purely of uncorrelated noise. Any remaining structured pattern would indicate that the autoencoder failed to fully model the underlying correlation dynamics. To assess this, we compute a row-averaged autocorrelation function (ACF) of the residual matrix. For each row of the residual, the mean is first removed and the autocorrelation is calculated up to a specified maximum lag. The resulting ACFs from all rows are then averaged to obtain a mean residual ACF. If the residual is consistent with uncorrelated stochastic fluctuations, the autocorrelation values for all non-zero lags should fluctuate around zero and remain within the expected statistical confidence bounds. We therefore compare the averaged ACF against the theoretical confidence limit for the uncorrelated fluctuations, $\pm z/\sqrt(N)$, where $N$ is the number of samples used to compute the autocorrelation for each row, and $z$ corresponds to the confidence level of 95\% ($z = 1.96$). When all non-zero lag values remain within this bound, the residual is considered statistically indistinguishable from uncorrelated noise, indicating that all significant structural signals are preserved in the denoised correlation function.

\subsection{Structural Similarity Index}

The Structural Similarity Index (SSIM)~\cite{wang2004ssim} is a widely used metric to compare two images based on luminance, contrast, and structural information, while emphasizing preservation of spatial features beyond simple pixel differences. The SSIM value ranges from $-1$ to $1$, where a value of 1 indicates identical images, values near zero indicate weak structural similarity, and negative values correspond to anticorrelated structures. SSIM provides a convenient way to assess how closely denoised \TTC reproduces the spatial patterns present in the raw data.

We computed the SSIM between \TTCraw and the corresponding denoised \TTCden for all samples in the validation dataset. 
The resulting SSIM values vary considerably in the validation dataset, indicating that FC-DAE reconstructs certain \TTC{} maps more reliably than others. Using a threshold of SSIM $\approx 0.15$ to distinguish meaningful structural recovery, about $24\%$ of the validation samples can be classified as reliably reconstructed.
It is important to note that a low SSIM value does not necessarily imply poor denoising performance. Since FC-DAE targets noise reduction, \TTCden should not strictly replicate the original \TTCraw pixels.
In cases where \TTCraw contains substantial noise, the removal of high-frequency fluctuations can reduce the apparent structural similarity between the two patterns. Therefore, SSIM should be interpreted primarily as an indicator of how well large-scale correlation patterns are preserved, rather than as a strict measure of denoising accuracy. 
For this reason, the SSIM analysis is complemented by additional diagnostics based on the statistical properties of the residual.

\section{Extracting Dynamic Parameters}
\label{sec:extract_dynam_params}
\begin{figure}
\centering
\includegraphics[scale=0.32]{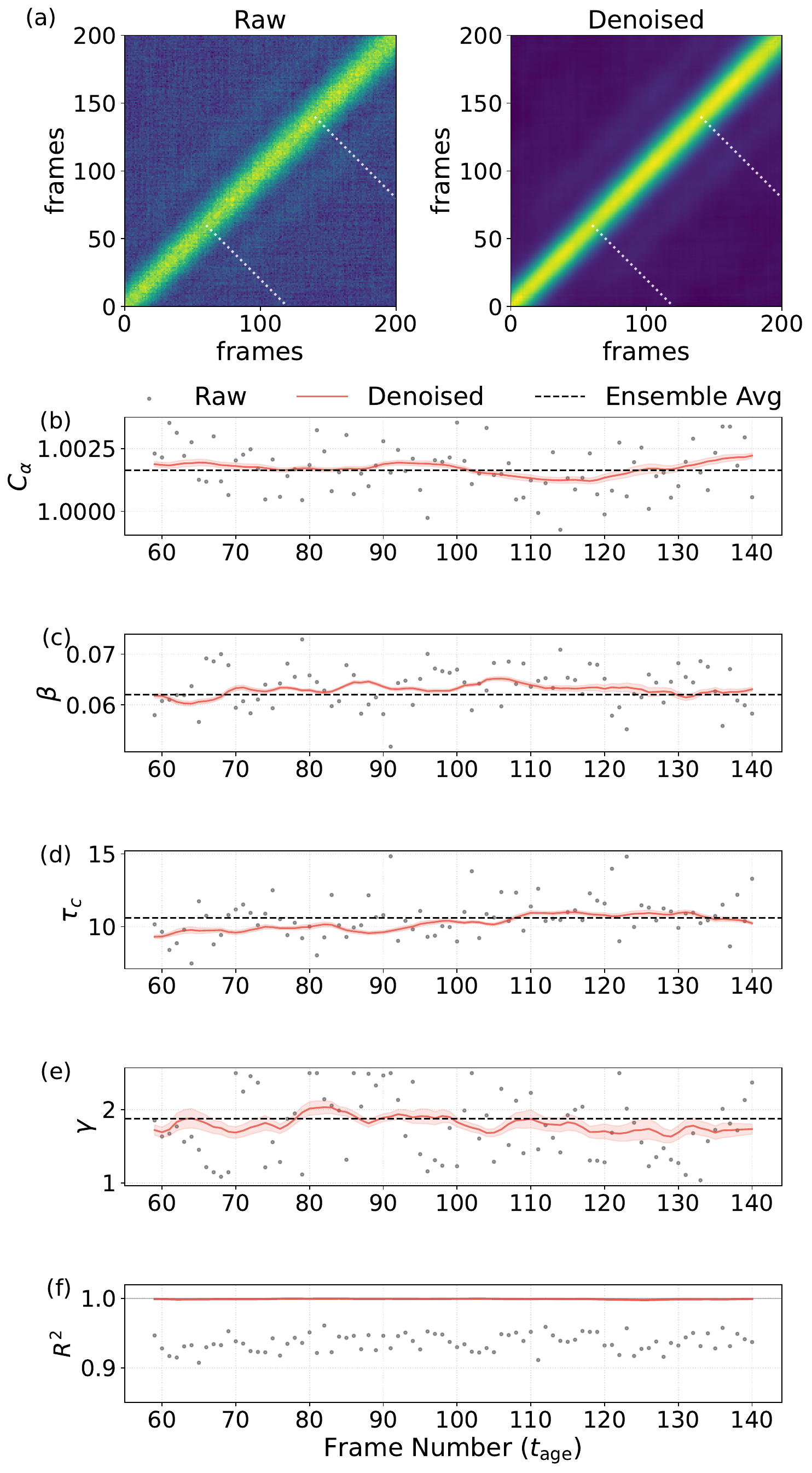}
\caption{Comparison of correlation data and extracted dynamic parameters. (a) Representative raw (left) and FC-DAE-denoised (right) \TTC data. Slices within dotted lines are used for fitting. (b–e) Evolution of the four primary KWW parameters ($\Cinf$, $\beta$, $\tau_c$, $\gamma$) as a function of frame number. The denoised results (red) exhibit substantially reduced statistical scatter compared to the raw slices (gray), while the ensemble average is shown as a dashed line. Light red shaded regions denote the $1\sigma$ fit uncertainties for the denoised results (raw uncertainties are omitted as they are substantially larger).
(f) The $R^2$ goodness-of-fit for each slice.}
\label{fig:Fit_param}
\end{figure}

To evaluate the impact of denoising on quantitative analysis, we examine the extraction of dynamic parameters from the strongly oscillatory \TTC{} data shown in the top row of Fig.~\ref{fig:Fit_param}. The corresponding \gTwo extracted from \TTCraw is not well described by a single KWW function. Instead, we employ a composite model consisting of a KWW term and a damped sinusoidal component to capture the pronounced peaking behavior observed in \gTwo. Both the raw and FC-DAE-denoised \TTC are sliced with a single-frame age window to obtain individual \gTwo curves, and each slice is independently fitted to extract the dynamic parameters. Due to the limited lag time, slices at the extreme edges of the \TTC are excluded from fitting.
Because the dynamics vary weakly along the diagonal of \TTC, we additionally collapse \TTC along the diagonal to form an ensemble-averaged \gTwo, which serves as a reference for the mean behavior. As shown in the lower rows (b-e) of Fig.~\ref{fig:Fit_param}, the parameters extracted from the denoised \TTC exhibit substantially reduced uncertainties and are significantly closer to the ensemble-averaged values than those obtained from the raw data, demonstrating that FC-DAE enables reliable parameter extraction even in this complex dynamical regime. The bottom row of Fig.~\ref{fig:Fit_param} displays the $R^2$ goodness-of-fit for each slice, demonstrating that the denoised data yields significantly better fits compared to the raw data.

\section{Recovery of Dynamics in Photon-Limited Regimes}
\label{sec:Reco-Dynamics}
To assess the ability of the FC-DAE to recover meaningful dynamics from noisy inputs, we employ two complementary approaches. First, we performed a controlled degradation study using bootstrapped datasets \TTC as described in Section~\ref{subsec:dat_processing} to systematically increase statistical noise. Second, we analyze \TTC obtained from regions of interest (ROIs) at increasing wavevector $q$, where reduced scattered intensity leads to lower SNR. 
Together, these approaches probe the robustness of the model across both synthetic and experimentally relevant noise conditions.

\subsection{Recovery under Controlled Noise Degradation}

\begin{figure}
\centering
\includegraphics[scale=0.3]{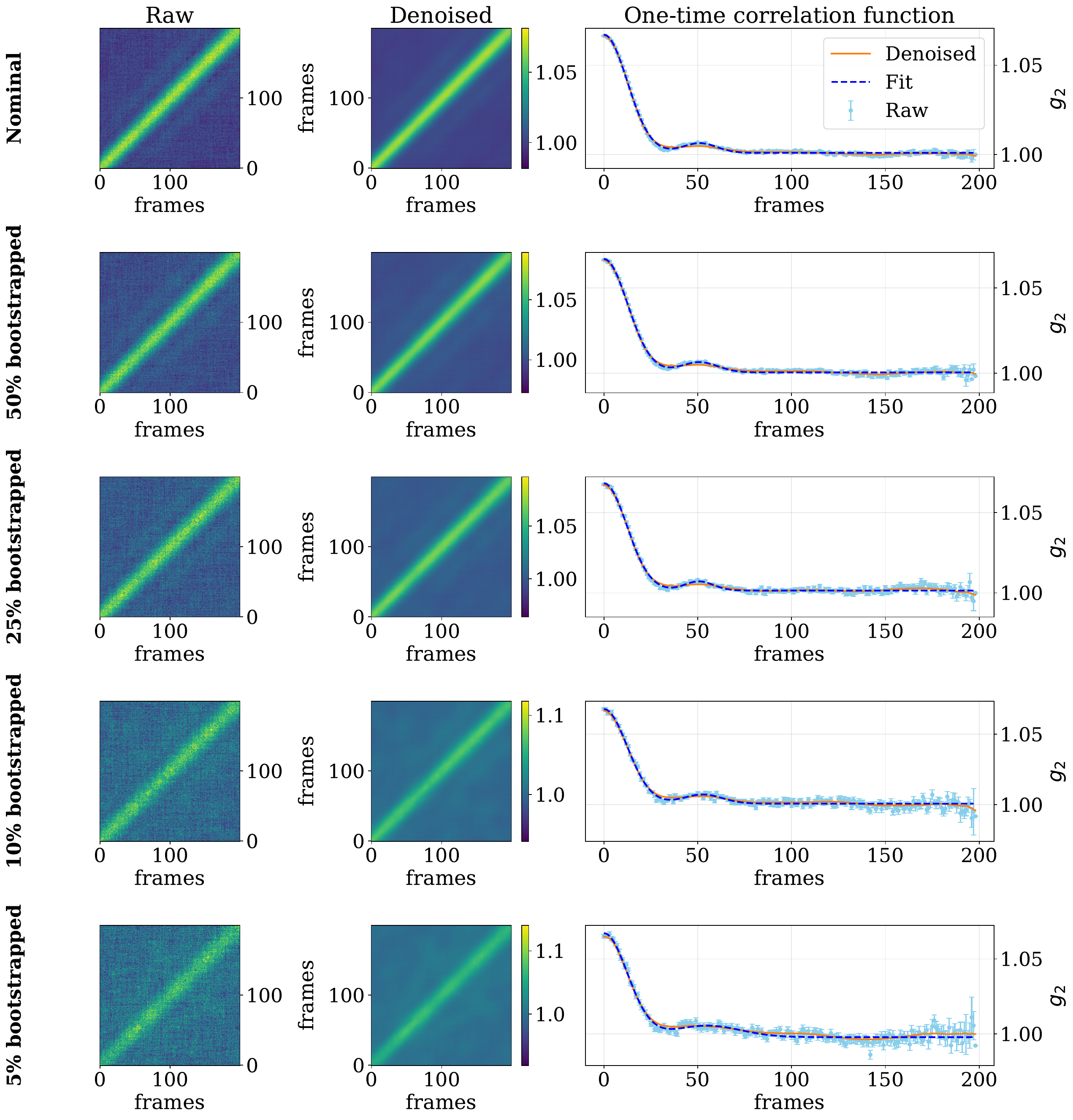}
\caption{Performance of the FC-DAE in recovering dynamics under progressive data degradation. Each row represents a different bootstrap fraction (Nominal, 50\%, 25\%, 10\%, and 5\%), corresponding to increasing noise levels. The first two columns show the raw input and the corresponding FC-DAE-denoised \TTC maps. The third column displays the extracted one-time correlation functions (\gTwo) for the raw (light blue dots) and denoised (orange line) data, along with a fit to the raw data (dashed blue line) using the function described in Section~\ref{sec:extract_dynam_params}. The FC-DAE demonstrates robust recovery of the oscillatory component across a wide range of noise levels.}

\label{fig:BS_study}
\end{figure}

Figure~\ref{fig:BS_study} illustrates the raw and FC-DAE-denoised \TTC maps together with the corresponding one-time correlation functions \gTwo. We focus on the recovery of the oscillatory (peaking) component in \gTwo, which is highly sensitive to noise and therefore provides a stringent test of reconstruction fidelity. The peak at approximately 50 frames is the focus of our evaluation. For moderate noise levels (50\%, 25\%, and 10\% bootstrapping), the FC-DAE reliably reconstructs both the peak position and amplitude, yielding denoised \gTwo curves that closely match the nominal behavior. 
At the most extreme degradation (5\%), the peak is strongly suppressed in the raw data due to loss of statistical contrast, and the reconstructed signal correspondingly shows reduced amplitude, indicating the practical limit of recoverable information. However, closer inspection of the raw \TTC at the 5\% level shows features in the vicinity of 160 frames (aging time) that are not present in the nominal case. Thus, the FC-DAE reconstruction is highlighting the consequence of spatially under-sampling the ensemble utilized for the \TTC computation. Use of root mean square between the raw nominal and 5\% {\TTC}s best demonstrate the dissimiliarity. In whole, these results demonstrate that the FC-DAE preserves nontrivial dynamical features over a wide range of noise levels while naturally reflecting that the provenance of inputs to the FC‑DAE is important for context when interpreting the outputs for materials discovery.

\subsection{Recovery across Wavevector-Dependent Noise Levels}

To quantify reconstruction performance across experimentally relevant conditions, we analyze \TTC at increasing wavevectors $q$. As $q$ increases, the scattered intensity per frame decreases, leading to progressively lower SNR. We consider \TTC datasets extracted from four regions of interest (ROIs) corresponding to consecutive $q$ values, $q = 1,\dots,4$. 
To quantify the SNR of \TTC, we adopt a definition tailored to XPCS data, where the one-time correlation function \gTwo\ typically exhibits a stretched-exponential decay rather than a well-defined peak. We estimate the SNR from an off-diagonal of the \TTC corresponding to a fixed lag time \lag, chosen such that the contrast decays to $\beta_0/e$, where $\beta_0$ is the initial contrast and \(e\) is Euler's number. This ensures that the SNR is evaluated in a region where the signal remains meaningful while avoiding trivial self-correlations. The SNR is then computed as

\begin{equation}
\mathrm{SNR} = \log_{10}\left( \frac{\langle \mathcal{D}_\tau \rangle}{\mathrm{Var}(\mathcal{D}_\tau)} \right),
\end{equation}

where $\mathcal{D}_\tau = \{ C_2(\tage, \tage + \tau) \mid \tage \}$ denote the set of values along this diagonal, i.e., at constant $\tau$ and varying age \tage. The $\langle \mathcal{D}_\tau \rangle$ and $\mathrm{Var}(\mathcal{D}_\tau)$ denote the mean and variance of the values along the selected diagonal.

\begin{figure}
\centering
\includegraphics[scale=0.3]{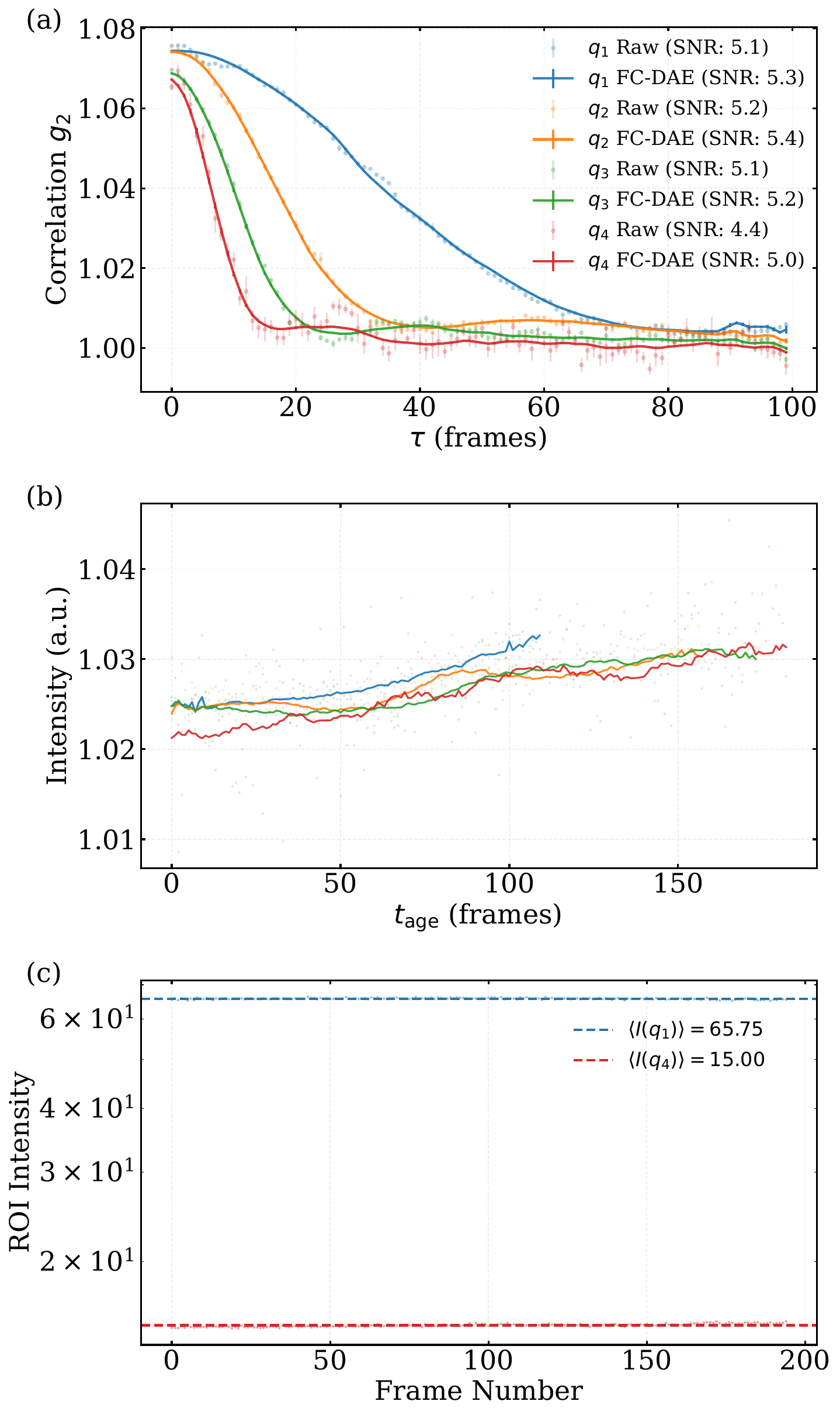}
\caption{Characterization of correlation functions and intensity stability across varying $q$-vectors.
    \textbf{(a)} Intensity autocorrelation functions, \gTwo, extracted from the central regions of \TTC matrices for $q_1, \dots, q_4$. Higher $q$ values correspond to lower photon flux, resulting in a diminished SNR in the raw data, which is subsequently recovered via FC-DAE denoising. 
    \textbf{(b)} Diagonal elements of the \TTC matrix at a constant \lag time, illustrating temporal intensity fluctuations and denoising performance. 
    \textbf{(c)} Average Region of Interest (ROI) intensities for $q_1$ and $q_4$. The disparity in mean counts highlights the challenge of resolving high-$q$ dynamics and demonstrates the potential for significant dose reduction enabled by the denoising algorithm.}
\label{fig:C2_atQs_study}
\end{figure}

Figure~\ref{fig:C2_atQs_study} summarizes the results. As shown in Fig.~\ref{fig:C2_atQs_study}(a), the SNR of \gTwo\ extracted from \TTCraw decreases with increasing $q$, while the FC-DAE significantly restores the correlation signal. Notably, the denoised \TTC at $q = 4$ achieves an SNR comparable to that of the raw \TTC at $q = 1$, despite substantially reduced photon statistics. Based on the average ROI intensities at $q = 1$ and $q = 4$, these results suggest that FC-DAE denoising could enable XPCS measurements of faster dynamics by increasing the achievable frame rate by a factor of $\sim4.4$, or equivalently allow measurements to be performed using only $\sim23\%$ of the original dose while maintaining comparable SNR. This interpretation is further supported by the diagonal analysis at constant \lag time shown in Fig.~\ref{fig:C2_atQs_study}(b). Fig.~\ref{fig:C2_atQs_study}(c) presents the average ROI intensities for $q = 1$ and $q = 4$, confirming that measurements at higher $q$ are performed under significantly lower intensity conditions.

These results demonstrate that the FC-DAE effectively compensates for reduced counting statistics, enabling reliable recovery of correlation signals at lower intensities. 
Consequently, denoising allows XPCS measurements to be performed at reduced flux or dose while maintaining comparable SNR, thereby extending the accessible experimental parameter space.

\section{Conclusions}

The proposed fully convolutional FC-DAE model provides robust denoising across a broad range of XPCS dynamics. Unlike latent-space DAE architectures, which require fixed-size inputs and highly specialized training domains, the FC-DAE can process \TTC inputs of arbitrary dimensions while generalizing across diverse dynamics and noise conditions.  This enhanced generalizability is fundamentally driven by the spatial invariance of the fully convolutional architecture; by utilizing localized smoothing kernels rather than coordinate-dependent fully connected layers, the network naturally scales the denoising strategies learned from local subregions across the entire correlation map. That said, the utilized training data, experimentally derived and of sufficient diversity and size, suggests that FC‑DAE can serve as a foundation for reproducible comparison of future denoising frameworks for \TTC data.
Visual and quantitative comparisons show that the FC-DAE preserves physically meaningful correlation structures more effectively than conventional DAE approaches, particularly for complex nonequilibrium dynamics where latent-space models tend to oversmooth or distort the signal. To assess reconstruction fidelity, reliability metrics are incorporated to identify cases where the denoised output may deviate from physically meaningful behavior, particularly for previously unseen or highly complex dynamical regimes.

Quantitative analysis further demonstrates that the FC-DAE substantially improves the effective SNR of the reconstructed \TTC, enabling reliable extraction of dynamic parameters under photon-limited conditions. Consequently, comparable resolution can be achieved using significantly smaller binning windows, leading to improved temporal resolution for evolving dynamics. The method also extends the accessible experimental parameter space by enabling analysis at larger $q$ values or reduced scattering intensities, where conventional approaches often fail because of poor counting statistics.

Overall, the FC-DAE provides a practical and computationally efficient framework for denoising XPCS correlation data while preserving underlying dynamics. By reducing the need for extensive temporal averaging, the approach enables access to rapidly evolving dynamics in XPCS experiments with weak scattering.


\section{Author Contributions}
T. Konstantinova, A. M. DeGennaro, A. M. Barbour, L. Wiegart, and N. Nellikunnummel conceived the idea. L. Wiegart and A. M. Barbour performed the beamline experiments and collected the XPCS data. L. Wiegart, A. M. Barbour, and T. Konstantinova selected individual datasets for model development. N. Nellikunnummel processed the data, and N. Nellikunnummel and T. Konstantinova developed the codebase. A. M. Barbour, L. Wiegart, and N. Nellikunnummel evaluated model performance. N. Nellikunnummel prepared the original draft of the manuscript, and all authors contributed to reviewing and editing the final document.

\section{Conflicts of interest}
There are no conflicts to declare.

\section{Data availability}

The source code used in this study is openly available on GitHub at https://github.com/NSLS2/Fully-Convolutional-DAE. The dataset used to train, test, and validate the model is available in the Zenodo repository at https://doi.org/10.5281/zenodo.20343306.

\section{Acknowledgment} 
We thank Andrei Fluerasu, Seher Karakuzu, Joshua Lynch, Max Rakitin, 
and Hiran Wijesinghe for insightful discussions regarding the model and its performance. This research used the CHX and CSX beamlines and resources of the National Synchrotron Light Source II, a U.S. Department of Energy (DOE) Office of Science User Facility operated for the DOE Office of Science by Brookhaven National Laboratory (BNL) under Contract No. DE-SC0012704 and BNL Laboratory Directed Research and Development (LDRD) Project No. 20-038, “Machine Learning for Real-Time Data Fidelity, Healing, and Analysis for Coherent X-ray Synchrotron Data.

\bibliographystyle{unsrt}
\bibliography{references}

@article{Madsen_2010,
doi = {10.1088/1367-2630/12/5/055001},
url = {https://doi.org/10.1088/1367-2630/12/5/055001},
year = {2010},
month = {may},
publisher = {},
volume = {12},
number = {5},
pages = {055001},
author = {Madsen, Anders and Leheny, Robert L and Guo, Hongyu and Sprung, Michael and Czakkel, Orsolya},
title = {Beyond simple exponential correlation functions and equilibrium dynamics in x-ray photon correlation spectroscopy},
journal = {New Journal of Physics}
}

@article{Shpyrko2014XPCS,
  title        = {X-ray photon correlation spectroscopy},
  author       = {Shpyrko, Oleg G.},
  journal      = {Journal of Synchrotron Radiation},
  year         = {2014},
  volume       = {21},
  number       = {5},
  pages        = {1057--1064},
  doi          = {10.1107/S1600577514018232},
  url          = {https://journals.iucr.org/s/issues/2014/05/00/vv5086/vv5086.pdf}
}

@book{GonzalezWoods2008,
  title     = {Digital Image Processing},
  author    = {Gonzalez, Rafael C. and Woods, Richard E.},
  edition   = {3rd},
  year      = {2008},
  publisher = {Pearson/Prentice Hall},
  address   = {Upper Saddle River, NJ, USA},
  isbn      = {978-0131687288}
}

@article{Al-amri2010NoiseRemoval,
  title        = {A Comparative Study of Removal Noise from Remote Sensing Image},
  author       = {Al-amri, Salem Saleh and Kalyankar, N. V. and Khamitkar, S. D.},
  journal      = {arXiv e-prints},
  year         = {2010},
  eprint       = {1002.1148},
  archivePrefix= {arXiv},
  primaryClass = {cs.CV},
  note         = {Mean filter and Gaussian filter among methods compared},
  url          = {https://arxiv.org/abs/1002.1148}
}

@article{Konstantinova2022SpeckleML,
  title        = {Machine learning for analysis of speckle dynamics: quantification and outlier detection},
  author       = {Konstantinova, Tatiana and Wiegart, Lutz and Rakitin, Maksim and DeGennaro, Anthony M. and Barbour, Andi M.},
  journal      = {Physical Review Research},
  volume       = {4},
  number       = {3},
  pages        = {033228},
  year         = {2022},
  doi          = {10.1103/PhysRevResearch.4.033228},
  url          = {https://doi.org/10.1103/PhysRevResearch.4.033228}
}

@article{Konstantinova2021,
  title        = {Noise reduction in X-ray photon correlation spectroscopy with convolutional neural networks encoder–decoder models},
  author       = {Konstantinova, Tatiana and Wiegart, Lutz and Rakitin, Maksim and DeGennaro, Anthony M. and Barbour, Andi M.},
  journal      = {Scientific Reports},
  volume       = {11},
  pages        = {14756},
  year         = {2021},
  doi          = {10.1038/s41598-021-93747-y},
  url          = {https://www.nature.com/articles/s41598-021-93747-y}
}

@article{Timmermann2022,
  author  = {Timmermann, Sonja and Starostin, Vladimir and Girelli, Anita and Ragulskaya, Anastasia and Rahmann, Hendrik and Reiser, Mario and Begam, Nafisa and Randolph, Lisa and Sprung, Michael and Westermeier, Fabian and Zhang, Fajun and Schreiber, Frank and Gutt, Christian},
  title   = {Automated matching of two-time X-ray photon correlation maps from phase-separating proteins with Cahn--Hilliard-type simulations using auto-encoder networks},
  journal = {Journal of Applied Crystallography},
  year    = {2022},
  volume  = {55},
  number  = {4},
  pages   = {751--757},
  doi     = {10.1107/S1600576722004435},
  url     = {https://doi.org/10.1107/S1600576722004435}
}

@article{Williams1970,
  title={Non-symmetrical dielectric relaxation behaviour arising from a simple empirical decay function},
  author={Williams, Graham and Watts, David C.},
  journal={Transactions of the Faraday Society},
  volume={66},
  pages={80--85},
  year={1970},
  publisher={Royal Society of Chemistry},
  doi={10.1039/TF9706600080}
}

@misc{kingma2017adammethodstochasticoptimization,
      title={Adam: A Method for Stochastic Optimization}, 
      author={Diederik P. Kingma and Jimmy Ba},
      year={2017},
      eprint={1412.6980},
      archivePrefix={arXiv},
      primaryClass={cs.LG},
      url={https://arxiv.org/abs/1412.6980}, 
}

@article{wang2004ssim,
  title={Image quality assessment: From error visibility to structural similarity},
  author={Wang, Zhou and Bovik, Alan C. and Sheikh, Hamid R. and Simoncelli, Eero P.},
  journal={IEEE Transactions on Image Processing},
  volume={13},
  number={4},
  pages={600--612},
  year={2004}
}

@article{Fluerasu2007Colloidal,
  title={Slow dynamics and aging in colloidal gels studied by x-ray photon correlation spectroscopy},
  author={Fluerasu, Andrei and Moussa{\"\i}d, Abdelhadi and Mariot, Alain and Petekidis, George},
  journal={Physical Review E},
  volume={76},
  number={1},
  pages={010401},
  year={2007},
  publisher={APS}
}

@article{hua2023discerning,
  title={Discerning element and site-specific fluctuations of the charge-orbital order in Fe 3 O 4 below the Verwey transition},
  author={Hua, Nelson and Li, Jianheng and Hrkac, Stjepan B and Barbour, Andi and Hu, Wen and Mazzoli, Claudio and Wilkins, Stuart and Kukreja, Roopali and Fullerton, Eric E and Shpyrko, Oleg G},
  journal={Physical Review Materials},
  volume={7},
  number={1},
  pages={014413},
  year={2023},
  publisher={APS}
}

@article{Burghardt2007Polymer,
  title={Equilibrium dynamics of a polymer bicontinuous microemulsion},
  author={Burghardt, Wesley R and Krishnan, K and Bates, Frank S and Lodge, Timothy P},
  journal={Macromolecules},
  volume={40},
  number={14},
  pages={5150--5158},
  year={2007},
  publisher={ACS Publications}
}

@article{Cipelletti2000Glassy,
  title={Universal aging features in the restructuring of fractal colloidal gels},
  author={Cipelletti, Luca and Manley, S and Ball, RC and Weitz, DA},
  journal={Physical Review Letters},
  volume={84},
  number={10},
  pages={2275},
  year={2000},
  publisher={APS}
}

@article{anthuparambil2023exploring,
  title={Exploring non-equilibrium processes and spatio-temporal scaling laws in heated egg yolk using coherent X-rays},
  author={Anthuparambil, Nimmi Das and Girelli, Anita and Timmermann, Sonja and Kowalski, Marvin and Akhundzadeh, Mohammad Sayed and Retzbach, Sebastian and Senft, Maximilian D and Dargasz, Michelle and Gutm{\"u}ller, Dennis and Hiremath, Anusha and others},
  journal={Nature Communications},
  volume={14},
  number={1},
  pages={5580},
  year={2023},
  publisher={Nature Publishing Group UK London}
}

@INPROCEEDINGS{FC-DAE_original,
  author={Kechris, Christodoulos and Delitzas, Alexandros and Matsoukas, Vasileios and Petrantonakis, Panagiotis C.},
  booktitle={2021 43rd Annual International Conference of the IEEE Engineering in Medicine \& Biology Society (EMBC)}, 
  title={Removing Noise from Extracellular Neural Recordings Using Fully Convolutional Denoising Autoencoders}, 
  year={2021},
  volume={},
  number={},
  pages={890-893},
  doi={10.1109/EMBC46164.2021.9630585}}

@article{sutton2008review,
  title={A review of X-ray intensity fluctuation spectroscopy},
  author={Sutton, Mark},
  journal={Comptes Rendus Physique},
  volume={9},
  number={5-6},
  pages={657--667},
  year={2008},
  publisher={Elsevier}
}

@article{liu2025dynamics,
  title={Dynamics of Thermally Driven Domain Transformation in Ferroelectric Thin Films},
  author={Liu, Rui and Gura, Anna and Sauyet, Theodore and Zhang, Yugang and Wiegart, Lutz and Fluerasu, Andrei and Dawber, Matthew},
  journal={Physical Review Letters},
  volume={134},
  number={5},
  pages={056801},
  year={2025},
  publisher={APS}
}

@article{zhang2018spatially,
  title={Spatially heterogeneous dynamics in a metallic glass forming liquid imaged by electron correlation microscopy},
  author={Zhang, Pei and Maldonis, Jason J and Liu, Ze and Schroers, Jan and Voyles, Paul M},
  journal={Nature communications},
  volume={9},
  number={1},
  pages={1--7},
  year={2018},
  publisher={Nature Publishing Group}
}

@article{spangenberg2021direct,
  title={Direct view on non-equilibrium heterogeneous dynamics in glassy nanorods},
  author={Spangenberg, Katharina and Hilke, Sven and Wilde, Gerhard and Peterlechner, Martin},
  journal={Advanced Functional Materials},
  volume={31},
  number={38},
  pages={2103742},
  year={2021},
  publisher={Wiley Online Library}
}

@article{huang2024momentum,
  title={Momentum transfer resolved electron correlation microscopy},
  author={Huang, Shuoyuan and Voyles, Paul M},
  journal={Ultramicroscopy},
  volume={256},
  pages={113886},
  year={2024},
  publisher={Elsevier}
}

@article{Horwath2024AINERD,
  author    = {Horwath, J. P. and Lin, X.-M. and He, H. and Zhang, Q. and Dufresne, E. M. and Chu, M. and others},
  title     = {AI-NERD: elucidation of relaxation dynamics beyond equilibrium through AI-informed X-ray photon correlation spectroscopy},
  journal   = {Nature Communications},
  volume    = {15},
  number    = {1},
  pages     = {5945},
  year      = {2024},
  publisher = {Nature Publishing Group},
  doi       = {10.1038/s41467-024-49381-z},
  url       = {https://nature.com}
}

\end{document}